\documentclass[sigconf]{acmart}

\copyrightyear{2019} 
\acmYear{2019} 
\acmConference[ACM-BCB '21]{12th ACM International Conference on Bioinformatics, Computational Biology and Health Informatics}{August 1--4, 2021}{}
\acmBooktitle{12th ACM International Conference on Bioinformatics, Computational Biology and Health Informatics (ACM-BCB '21), August 1--4, 2021}
\acmDOI{10.1145/3307339.3343250}
\acmISBN{978-1-4503-6666-3/19/09}

\usepackage{booktabs} 
\usepackage{hyperref}
\usepackage{url}
\usepackage{multirow}
\usepackage{multicol}
\usepackage{caption}
\usepackage{flushend,balance}
\usepackage{setspace}
\usepackage{tikz,adjustbox}
\usetikzlibrary{intersections}
\usepackage{booktabs}
\usetikzlibrary{calc}
\usetikzlibrary{decorations.pathreplacing}
\usetikzlibrary{shapes,arrows,positioning}
\usetikzlibrary{chains, positioning}
\tikzset{
  treenode/.style = {shape=rectangle, rounded corners,
                     draw, align=center,
                     top color=white, bottom color=blue!20},
  root/.style     = {treenode, font=\Large, bottom color=red!30},
  env/.style      = {treenode, font=\ttfamily\normalsize},
  dummy/.style    = {circle,draw}
}

\usepackage{pgfplots}
\pgfplotsset{compat=1.12}
\usepackage{graphicx}
\usepackage{amsmath,amsthm,bbm,mathtools,adjustbox}
\usepackage{algorithm,algorithmicx,algpseudocode}
\usepackage{algorithmicx,algorithm}
\usepackage{enumitem}
\usepackage{graphicx}
\usepackage{float,color}
\usepackage{xspace}

\usepackage{amsmath,amsfonts,bm}

\def\eqref#1{equation~\ref{#1}}

\def\1{\bm{1}}

\DeclareMathAlphabet{\mathsfit}{\encodingdefault}{\sfdefault}{m}{sl}
\SetMathAlphabet{\mathsfit}{bold}{\encodingdefault}{\sfdefault}{bx}{n}

\usepackage{subfigure}

\theoremstyle{definition}

\newtheorem{problem}{Problem}

\newcommand{\mname}{\texttt{DrugCLIP}\xspace }

\newcommand{\bfe}{\mathbf{e}}

\newcommand{\bfh}{\mathbf{h}}

\newcommand{\bfm}{\mathbf{m}}

\newcommand{\RB}{\mathbb{R}}

\newcommand{\calN}{\mathcal{N}}

\newcommand{\Epsilon}{E}

\newcommand*{\origrightarrow}{}

\renewcommand*{\textrightarrow}{\fontfamily{cmr}\selectfont\origrightarrow}

\begin{document} 

\fancyhead{}

\title{\mname: Contrastive Drug-Disease Interaction For Drug Repurposing}

\author{Yingzhou Lu}
\authornote{Both authors contributed equally to this research.}
\email{lyz66@stanford.edu}
\affiliation{%
  \institution{Stanford University}
  \city{Stanford, CA}
  \country{USA}
}

\author{Yaojun Hu}
\authornote{Both authors contributed equally to this research.}
\email{yaojunhu@zju.edu.cn}
\affiliation{%
  \institution{Zhejiang University}
  \city{Hangzhou}
  \country{China}
}

\author{Chenhao Li}
\email{cl89@illinois.edu}
\affiliation{%
  \institution{University of Illinois Urbana-Champaign}
  \city{Champaign, IL}
  \country{USA}
}

\begin{abstract}
Bringing a novel drug from the original idea to market typically requires more than ten years and billions of dollars. 
To alleviate the heavy burden, a natural idea is to reuse the approved drug to treat new diseases. The process is also known as \textit{drug repurposing} or \textit{drug repositioning}. Machine learning methods exhibited huge potential in automating drug repurposing. However, it still encounter some challenges, such as lack of labels and multimodal feature representation. To address these issues, we design DrugCLIP, a cutting-edge contrastive learning method, to learn drug and disease's interaction without negative labels. 
Additionally, we have curated a drug repurposing dataset based on real-world clinical trial records. Thorough empirical studies are conducted to validate the effectiveness of the proposed DrugCLIP method. 
\end{abstract}



\maketitle

\section{Introduction}

\begin{figure}[ht]
\centering
\includegraphics[width = 0.62\linewidth]{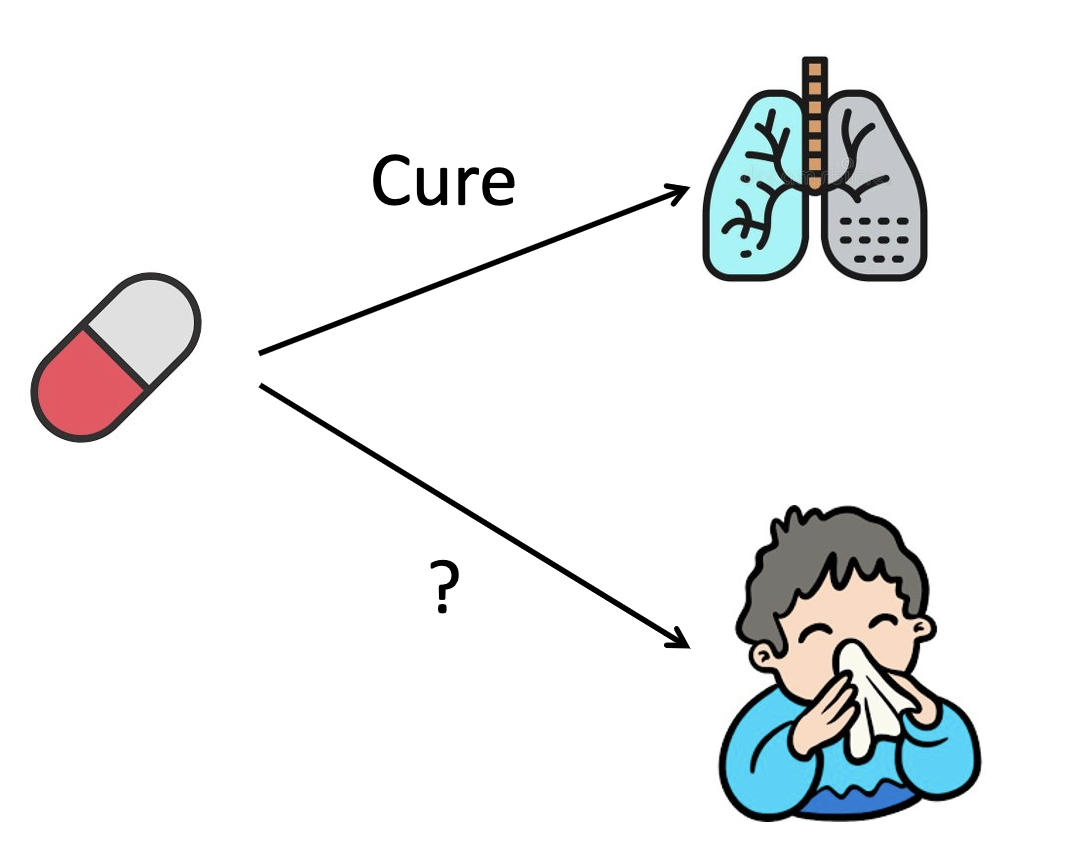}
\caption{
An example of drug repurposing: the drug that can treat lung disease can be reused to treat colds. 
Drug repurposing involves using approved drug candidates (or drug candidates that pass a Phase I trial) to treat new diseases. 
Compared with \textit{de novo} drug design that designs drug molecules from scratch, drug repurposing saves many resources and time because the safety issue of the drug has been largely alleviated. 
}
\label{fig:repurposing2}
\end{figure}

Bringing a novel drug from the original idea to market typically requires more than ten years and billions of dollars. 
Also, the safety issue is a major failure reason in drug discovery and development~\cite{bohacek1996art}. 
If the drug candidate fails to pass Phase I due to safety issues (e.g., the drug candidate is toxic to the human body), all the investment from early drug discovery to Phase I would yield nothing. 
To alleviate the heavy burden, a natural idea is to reuse the approved drug to treat new diseases. The process is also known as drug repurposing or drug repositioning~\cite{xu2020electronic}, as illustrated in Figure~\ref{fig:repurposing2}. 
The repurposed drugs typically do not have safety concerns and can directly enter phases II and III to evaluate their efficacy in treating certain diseases, which can save lots of resources, time, and funding on discovering new drugs.

The power of AI methods, especially machine learning (ML) technologies, has vastly increased since the early 2010s because of the availability of large datasets, coupled with new algorithmic techniques and aided by fast and massively parallel computing power. 
Machine learning-assisted drug repurposing has been extensively studied by formulating it into a drug-target interaction prediction problem, where the target protein is usually closely associated with human diseases. 
For example, DeepPurpose~\cite{huang2020deeppurpose} investigates 15 neural network architectures representing drug molecules and target proteins. 
The empirical results demonstrate the high accuracy of the drug-target interaction. 
Specifically, DeepPurpose can identify the ideal drug that could be validated by real-world evidence for some well-known indications, e.g., COVID-19.

\paragraph{Challenges.} However, existing drug repurposing methods still face a couple of challenges. 
\begin{itemize}
\item C1: lack of high-quality data. Clinical trial data are usually highly noisy and sensitive and raise privacy issues, which hinders AI's deployment. 
\item C2: lack of labeled data. Drug repurposing lacks labeled data, especially the negative samples. That is to say, we only know a certain drug could treat a certain disease and do not know a certain drug cannot treat a certain disease. 
\item C3: multimodal data interaction. In drug repurposing, there are multimodal data, such as drug molecules and disease codes. It is challenging to integrate these multimodal data in an end-to-end learning framework. 
\end{itemize}

\paragraph{Solutions.} To address these challenges, we propose the following solutions. 
\begin{itemize}
\item data curation. To address the first challenge (C1), we curate a high-quality drug repurposing dataset that are collected from multiple public data sources. 
\item constrative learning. To address the second challenge (C2), we design a contrastive drug-disease interaction framework that could produce negative samples and discriminate positive ones from negative ones. 
\item multimodal data representation. To address the third challenge (C3), we design representation learning methods for drug molecules and disease codes. 
\end{itemize}

\paragraph{Main Contributions. } For ease of exposition, the major contribution of this manuscript can be summarized as follows. 
\begin{itemize}
\item \textbf{Data.} We curate a ready-to-use dataset specialized for drug repurposing. The dataset comes from clinical trial data and contains around 35K clinical trials spanning from the early 2000s to the present.  
\item \textbf{Method.} We develop DrugCLIP, a state-of-the-art deep contrastive learning method (CLIP) tailored to drug repurposing, which models the interaction between drug molecules and disease codes. 
\item \textbf{Results.} We conduct extensive experiments to validate the effectiveness of the proposed method. Specifically, the proposed methods achieve 16.5\% improvement in hit rate over the best baseline method. 
\end{itemize}

The rest of the paper is organized as follows. 
First, Section~\ref{sec:related} briefly reviews the related works in using AI for drug discovery and development tasks. Then, we elaborate on our method in Section~\ref{sec:method}. After that, empirical studies are described in Section~\ref{sec:experiment}. Finally, we conclude the paper in Section~\ref{sec:conclusion}.

\section{Related Works}
\label{sec:related}

\paragraph{AI for Drug Discovery. }
Drugs are essentially molecular structures with desirable pharmaceutical properties. The goal of \textit{de novo} drug design is to produce novel and desirable molecule structures from scratch. The word ``\textit{de novo}'' means from the beginning. 
The whole molecule space is around $10^{60}$~\cite{bohacek1996art,huang2022artificial,huang2020deeppurpose}. Most of the existing methods rely heavily on brute-force enumeration and are computationally prohibitive. 
Generative models are able to learn the distribution of drug molecules from the existing drug database and then draw novel samples (i.e., drug molecules) from the learned molecule distribution, including variational autoencoder (VAE)~\cite{gomez2018automatic,zhang2021ddn2,jin2018junction}, generative adversarial network (GAN)~\cite{de2018molgan}, energy-based model (EBM)~\cite{liu2021graphebm,fu2022antibody,fu2021differentiable}, diffusion model~\cite{xu2021geodiff}, reinforcement learning (RL)~\cite{olivecrona2017molecular,zhou2019optimization,fu2022reinforced}, genetic algorithm~\cite{jensen2019graph}, sampling-based methods~\cite{fu2021mimosa,fu2022sipf}, etc.

\paragraph{AI for Drug Development (Clinical Trial)}
AI, especially deep learning methods, has great potential in aiding many clinical trial problems. 
Specifically, \cite{zhang2020deepenroll,gao2020compose} predict patient-trial matching based on trial eligibility criteria and patient electronic health records (EHR); \cite{wu-genetic-lesion} utilized support vector machines (SVMs) to forecast genetic lesions based on cancer clinical trial documents. \cite{rajpurkar-gbdt} utilized gradient-boosted decision trees (GBDTs) to predict improvements in symptom scores by integrating treatment symptom scores and EEG measures in the context of antidepressant treatment for depressive symptoms. Meanwhile, \cite{hong-drug-toxicity} directed their efforts towards projecting clinical drug toxicity based on features related to drug properties and target properties, employing an ensemble classifier consisting of weighted least squares support vector regression. Machine learning can also play a pivotal role in generating simulated data to identify more efficient statistical outcome measures \citep{sangari-ml-for-breast-cancer}. One study proposes that employing an AI algorithm to forecast individual patient outcomes and pinpoint those likely to progress rapidly, leading to earlier trial endpoints, could result in shorter trial duration~\cite{lee-ai-for-trials}. \cite{fu2022hint,lu2024uncertainty,chen2024uncertainty} propose to predict clinical trial approval based on drug molecule structure, disease code, and trial eligibility criteria. 
\cite{wang2024twin} design a TWIN-GPT model by finetuning standard GPT model to synthesize patient visit history to mimic clinical trials and predict trial outcomes. 
\cite{yue2024trialdura} predicts clinical trial duration using textual information of various trial features (disease, drug, eligibility criteria) with a pretrained BioBERT model as a text feature enhancement.

\section{Method}
\label{sec:method}

\subsection{Problem Formulation}
The goal of drug repurposing is to reuse the approved drug to treat new diseases. It is formulated as the drug-disease interaction problem, which is a binary classification task. 
\begin{problem}[Drug-disease interaction]
\label{problem:drug_disease_interaction}
Suppose we are given a set of disease codes $d$ to represent the disease and treatment (usually one drug molecules, denoted $m$), the goal of drug-disease interaction is to forecast if the drug could treat these diseases. 
It is formally defined as 
\begin{equation}
y = f_{\bm{\theta}}(d, m),     
\end{equation}
where $y \in \{0,1\}$ is the binary label, indicating whether the treatment can treat the disease. 
\end{problem}
Both drug molecules and disease codes are structured data and are not ready for machine learning like a feature vector. To proceed, we first utilize representation learning to transform drug molecules and disease codes into meaningful representation vectors (i.e., embedding) and then perform binary classification based on the embeddings. 
Thus, the whole pipeline is divided into three modules: (i) drug molecule representation, (ii) disease code representation, and (iii) contrastive drug-disease interaction learning, which will be elaborated as follows. 
For ease of exposition, we list all the mathematical notations in Table~\ref{table:notation}.

\subsection{Drug Molecule Representation} 
\label{sec:drug}
Compared with the SMILES string, the molecular graph is a more expressive and intuitive data representation of a molecule. 
Each node corresponds to an atom in the molecule, while an edge corresponds to a chemical bond. 
The molecular graph mainly contains two essential components: node identities and node interconnectivity~\cite{coley2017convolutional}. 

The nodes' identities include atom types, e.g., carbon, oxygen, nitrogen, etc. 
The nodes' connectivity can be represented as an adjacency matrix, where the (i,j)-th element denotes the connectivity between $i$-th and $j$-th nodes. Unlike the binary adjacency matrix that only reveals the connectivity, the bond type is an important molecule ingredient. 
To incorporate the bond type information, we use 0 to represent the non-connectivity, 1 to represent the single bond, 2 for the double bond, 3 for the triple bond, and 4 for the aromatic bond.

\begin{figure}
 \centering
 \includegraphics[width=\linewidth]{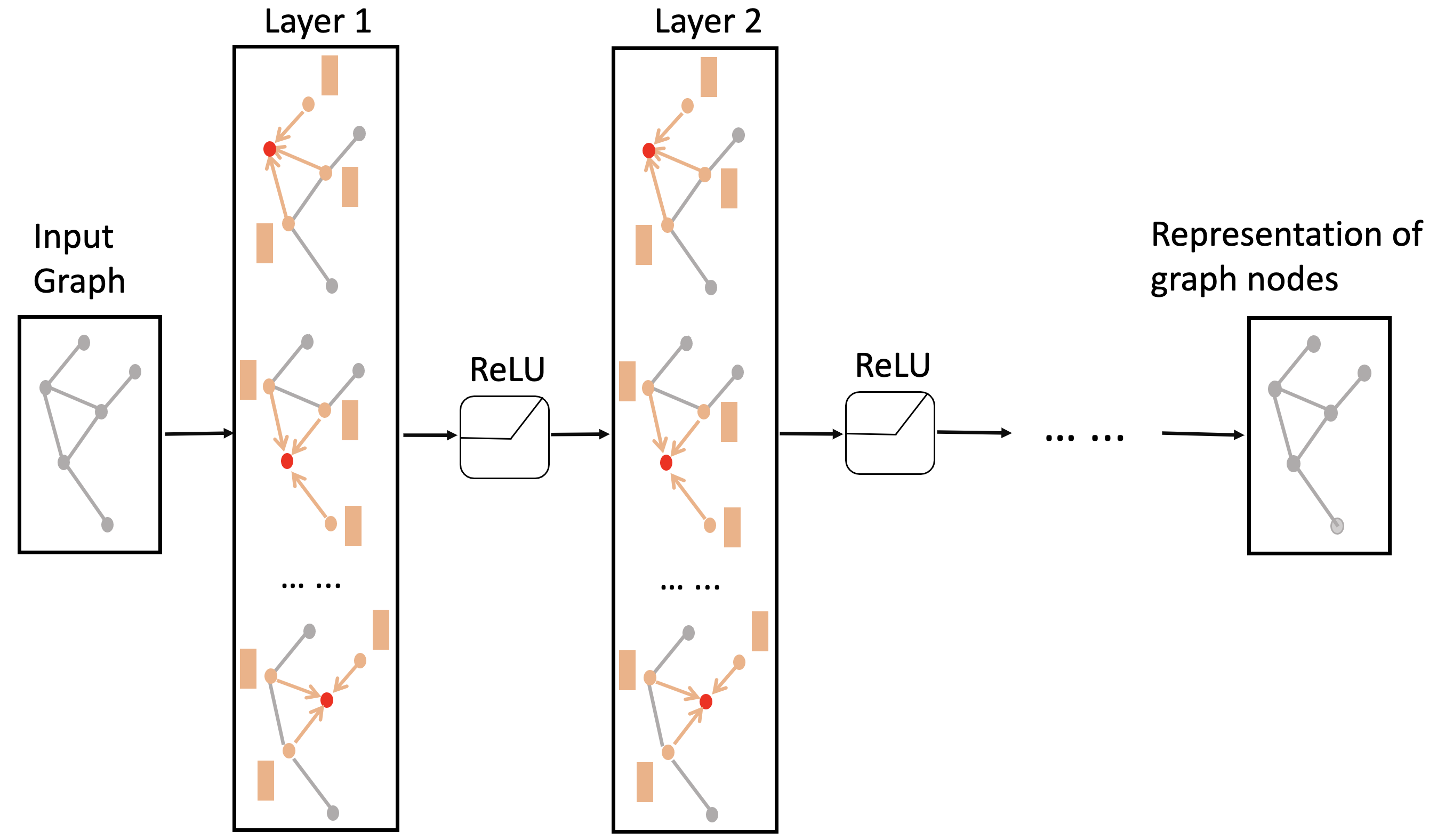}
 \caption{Message passing neural network for drug molecule representation. For each node (red) in the input molecular graph, message passing neural network iteratively updates its representation by aggregating representations of its neighbors (orange). }
 \label{fig:gnn}
\end{figure}

Drug representation aims to represent drug molecules in the embedding vectors. 
Many deep neural network models can be leveraged to achieve this goal, including graph neural networks, and recurrent neural networks. 
Formally, Message Passing Neural Network (MPNN)~\cite{dai2016discriminative} updates the information of edges in a graph. 
First, on the node level, each node $v$ has a feature vector denoted $\bfe_v$. 
For example, node $v$ in a molecular graph $G$ is an atom, $\bfe_v$ includes the atom type, valence, and other atomic properties. 
$\bfe_v$ can be a one-hot vector indicating the category of the node $v$. 
On the edge level, $\bfe_{uv}$ is the feature vector for edge $(u,v) \in \Epsilon$. 
$\calN(u)$ represents the set of all the neighbor nodes of the node $u$. 
At the $l$-th layer, $\bfm^{(l)}_{uv}$ and $\bfm^{(l)}_{vu}$ are the directional edge embeddings representing the message from node $u$ to node $v$ and vice versa.  
They are iteratively updated as
\begin{equation}
\begin{aligned}
\bfm_{uv}^{(l)} = f_1\bigg( \bfe_{u} \oplus \bfe_{uv}^{(l-1)} \oplus \sum_{w \in \calN(u)\backslash v} \bfm_{wu}^{(l-1)} \bigg),\ \ \ \ l = 1,\cdots, L, 
\end{aligned}
\end{equation} 
where $\oplus$ denotes the concatenation of two vectors; $f_1(\cdot)$ is a multiple layer perceptron (MLP), $\bfm_{uv}^{(l)}$ is the message vector from node $u$ to node $v$ at the $l$-th iteration, whose initialization is all-0 vector, i.e., $\bfm_{uv}^{(0)} = \mathbf{0}$, following the rule of thumb~\cite{fu2020core,fu2021mimosa}. 
After $L$ steps of iteration ($L$ is the depth), another multiple layer perceptron (MLP) $f_2(\cdot)$ is used to aggregate these messages. Each node has an embedding vector as
\begin{equation}
\begin{aligned}
\bfh_{u} = f_2\bigg(\bfe_u \oplus \sum_{v \in {\calN(u)}} \bfm_{vu}^{(L)} \bigg).
\end{aligned}
\end{equation}
We are interested in the representation of the entire graph, the READOUT function can aggregate all the nodes' embeddings in the last layer to yield graph-level representation, 
\begin{equation}
\label{eqn:readout}
\bfh_{G} = \text{READOUT}\bigg(\big\{ \bfh_v^{(L)} \big\}_{v\in V}\bigg),
\end{equation}
where the READOUT function applies over all nodes $V$ in graph $G$ (e.g., it can be an average or a summation function over all nodes).
Figure~\ref{fig:gnn} illustrates how messages are transmitted between the neighborhood nodes. 
There are several variations of graph neural networks, including representative ones, that will be discussed.

\begin{table*}[]
\centering
\caption{Mathematical notations and their explanations. The first block includes the notations for drug molecular representation (Section~\ref{sec:drug}), while the second and third blocks are for the notation for disease code (Section~\ref{sec:disease_code}) and DrugCLIP (Section~\ref{sec:drugclip}), respectively.  }
\vspace{2mm}
\resizebox{1.94\columnwidth}{!}{
\begin{tabular}{c|p{1.65\columnwidth}}
\toprule[1pt]
Notations & Explanations \\ 
\hline 
$G = (V,E)$ & graph structure of drug molecule. \\ 
$V$ & the set of nodes in the molecular graph $G$. \\
$E$ & the set of edges in the molecular graph $G$. \\ 
$L$ & number of layers in graph neural network, also known as the depth of the graph neural network. \\
$d$ & dimension of embedding. \\ 
$\bfh^{(l)}_u \in \RB^{d}$ & node embedding of the node $u$ at the $l$-th layer, $l=1,\cdots,L$ \\ 
$\mathcal{N}(v)$ & the set of all the neighbor nodes of the node $v$ \\
READOUT & readout function converts a set of node embeddings into graph-level embedding, which can be summation or average function.  \\ 
$\bfh_G$ & graph-level representation.  \\ 
$\bfe_v$ & input feature of the node $v$ \\ 
$\bfe_{uv}$ & input feature of the edge $(u,v)$ that connects nodes $u$ and $v$ \\ 
$\bfm^{(l)}_{uv}$ & edge embedding that represents the message from node $u$ to node $v$ at the $l$-th layer, which is directional.   \\ 
$f_1(\cdot), f_2(\cdot)$ & multiple layer perceptron (MLP) \\ 
$\oplus$ & the concatenation of two vectors \\ 
\hline 
 $\alpha$ & attention weight after normalization (softmax) \\ 
 $\bfe_i$ & basic embedding of the disease code $i$ \\ 
 $\bfh_i$ & GRAM representation of disease code $i$ \\ 
 $\alpha_{ji}$ & attention weight of disease code $i$ and its ancestor code $j$ \\ 
 $\phi(\cdot)$ & MLP used in GRAM \\ 
 $\text{Ancestors}(i)$ & the set of all the ancestor codes of the disease code $i$. \\ 
\hline 
$s_{i,j}$ & interaction between the $i$-th drug and the $j$-th disease code, i.e., $(i,j)$-th element in similarity matrix \\
$y_{i,j}$ & binary variable that indicates whether the $i$-th drug could treat the $j$-th disease code in historical clinical trial.  \\ 
$\sigma(\cdot): \mathbb{R} \xrightarrow[]{} \mathbb{R}$ & sigmoid function \\ 
$\mathcal{L}$ & the whole learning objective, i.e., loss function \\ 
\bottomrule[1pt]
\end{tabular}}
\label{table:notation}
\end{table*}

\subsection{Disease Code Representation}
\label{sec:disease_code}

Diseases lie at the heart of drug discovery. 
There are several standardized disease coding systems that healthcare providers use for the electronic exchange of clinical health information, 
including the International Classification of Diseases, Tenth Revision, Clinical Modification (ICD-10-CM), 
The International Classification of Diseases, Ninth Revision, Clinical Modification (ICD-9-CM), 
and Systematized Nomenclature of Medicine -- Clinical Terms (SNOMED CT)~\cite{anker2016welcome}.

These coding systems contain disease concepts organized into hierarchies. 
We take the ICD-10-CM code as an example. 
ICD-10-CM is a seven-character, alphanumeric code. 
Each code begins with a letter, and two numbers follow that letter. 
The first three characters of ICD-10-CM are the ``category''. The category describes the general type of injury or disease. 
A decimal point and the subcategory follow the category. 
For example, the code ``G44'' represents ``Other headache syndromes''; 
the code ``G44.31`` represents ``Acute post-traumatic headache'';  
the code ``G44.311'' represents ``Acute post-traumatic headache, intractable''. G44.311 has two ancestors: G44 and G44.31, where an ancestor represents a higher-level category of the current code~\cite{lu2021cot,lu2019integrated,wu2022cosbin}.   
The description of all the ICD-10-CM codes is available at \url{https://www.icd10data.com/ICD10CM/Codes}.

The disease information comes from its description and its corresponding ontology, such as disease hierarchies like the International Classification of Diseases (ICD)~\cite{anker2016welcome}. 
An ancestor of code represents a higher-level category of the current code. 
For example, in ICD 10 code~\cite{anker2016welcome,chen2021data}, ``D41'' (urinary organs neoplasm) and ``D41.2'' (ureter neoplasm) are the ancestors of  ``D41.20'' (right ureter neoplasm). 
``C34'' (malignant neoplasm of bronchus and lung) and ``C34.9'' (malignant neoplasm of unspecified part of bronchus or lung) are the ancestors of ``C34.91'' (malignant neoplasm of unspecified part of right bronchus or lung)~\cite{lu2018multi}.

Graph-based attention model (GRAM)~\cite{choi2017gram} designs an attention-based graph model to leverage the hierarchical information inherent to medical ontologies. 
Specifically, each disease code is assigned a basic embedding, e.g., the disease code $d_i$ has basic embedding, denoted $\bfe_i \in\RB^{d}$. 
Then, to impute the hierarchical dependencies, the embedding of current disease $d_i$ (denoted $\bfh_i$) is represented as a weighted average of the basic embeddings ($\bfe\in \RB^{d}$) of itself and its ancestors, the weight is evaluated by the attention model. 
It is formally defined as 
\begin{equation}
\label{eqn:gram}
\bfh_i = \sum_{j\in \text{Ancestors}(i)\cup \{i\}} \alpha_{ij} \bfe_j, 
\end{equation}
where $\alpha_{ji}\in (0,1)$ represents the attention weight and is defined as

\begin{figure}[t]
\centering
\includegraphics[width = 0.97\linewidth]{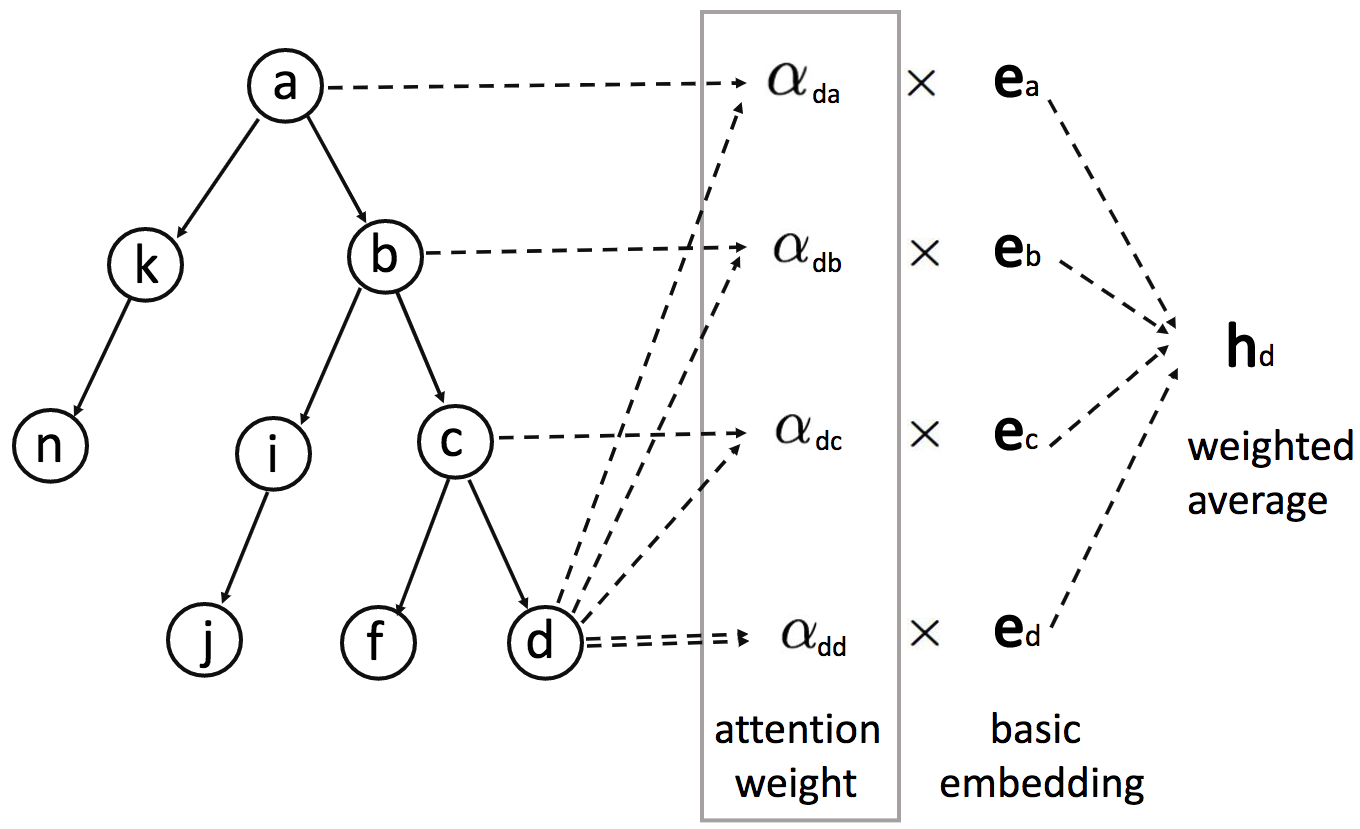}
\caption{Illustration of Graph-based attention model (GRAM), where the representation of the disease code is a weighted average of itself and all of its ancestors, and the weight is evaluated by attention mechanism. }
\label{fig:gram}
\end{figure}

\begin{equation}
\label{eqn:gram_attention}
\begin{aligned}
& \alpha_{ji} = \frac{\exp\big(\phi([\bfe_j^\top,\bfe_i^\top]^\top)\big)}{\sum_{k\in \text{Ancestors}(i)\cup \{i\}} \exp\big(\phi([\bfe_{k}^\top,\bfe_i^\top]^\top)\big)},\\ 
& \sum_{j\in \text{Ancestors}(i)\cup \{i\}} \alpha_{ji} = 1,
\end{aligned}
\end{equation}
where the attention model $\phi(\cdot)$ is an MLP with a single hidden layer, the input is the concatenation of the basic embedding, the output is a scalar, $\bfe_i$ serves as the query while all the ancestors embeddings $\big\{\bfe_j \big\}$ serve as the keys. 
$\text{Ancestors}(i)$ represents the set of all the ancestors of the disease code $d_i$.

\subsection{DrugCLIP: Contrastive Drug-Disease Interaction}
\label{sec:drugclip}

Recent advancements in large-scale image-text pre-training models, such as CLIP~\cite{radford2021learning}, have achieved significant achievements in both computer vision and natural language processing fields. CLIP relies on predicting the accurate pairing of images and corresponding textual descriptions within a training batch. This simultaneous learning of image and text representations, executed on extensive datasets of image-text pairs, yields versatile representations that are readily applicable to a wide array of subsequent tasks. Drawing parallels from CLIP's success, we assume that the synergistic knowledge extracted from drug and disease holds substantial promise for enhancing performance on drug repurposing.

Formally, suppose we have a drug database (denoted $\{m_1, m_2, \cdots, \}$) and a disease code database (denoted $\{d_1, d_2, \cdots, \}$). Then, we construct similarity matrix based on their representation. Specifically, the interaction between the $i$-th drug and the $j$-th disease code corresponds to $(i,j)$-th element, denoted $s_{i,j}$. It is defined as the cosine similarity between their representations. 
\begin{equation}
{s}_{i,j} = \frac{\bfh_{m_i}^\top \cdot \bfh_{d_j}}{||\bfh_{m_i}||_2 \cdot ||\bfh_{d_j}||_2}. 
\end{equation}
Then, we define the groundtruth to provide the supervision, which is denoted $y$. Specifically, assume the $i$-th drug could treat the $j$-th disease code in historical clinical trial, then we let $y_{i,j} = 1$. Otherwise, we let $y_{i,j} = 0$. 
The whole learning objective is to minimize the cross entropy loss as 
\begin{equation}
\label{eqn:loss}
\mathcal{L} = - \sum_{i,j} y_{i,j} \log \sigma(s_{i,j}) + (1 - y_{i,j}) (\log (1 -\sigma(s_{i,j}))), 
\end{equation}
where $\sigma(\cdot): \mathbb{R} \xrightarrow[]{} \mathbb{R}$ is the sigmoid function, which is used to produce the probability. That is, $\sigma(s_{i,j})$ denotes the probability that $i$-th drug could treat the $j$-th disease code.

\section{Experiment}
\label{sec:experiment}

\begin{table*}[h!]
\footnotesize
\centering
\caption{Size of selected clinical trials for various time bins. }
\label{table:statistics}
\resizebox{1.64\columnwidth}{!}{
\begin{tabular}{lccccccccc}
\toprule[0.6pt]
Dataset & 2000-2002 & 2003-2005 & 2006-2008 & 2009-2011 & 2012-2014 & 2015-2017 & 2018-2020 & 2021-2023 \\ \hline 
\# Trials & 1,490 & 3,791 & 5,065 & 5,539 & 5,388 & 5,254 & 5,891 & 3,764 \\ 
\bottomrule[0.6pt]
\end{tabular}}
\end{table*}

In this section, we demonstrate the experimental results. We start with the experimental setup, including the data curation process (Section~\ref{sec:data}), baseline methods (Section~\ref{sec:baseline}), and evaluation metrics (Section~\ref{sec:metric}). 
Then, we present the quantitative results in Section~\ref{sec:result}. 

\subsection{Data} 
\label{sec:data}
We collect the data from the following public sources.
\begin{itemize}
\item \textbf{Clinical Trial Records} (\url{http://clinicaltrials.gov}) includes more than 420K registered clinical trials. Each record describes a clinical trial in detail. Our experiment includes trials that are (1)  interventional, (2) tests a single drug, (3) the tested drug is a small molecule drug, and (4) the drug molecule is available in DrugBank. Note that in clinical trial data, diseases are described using unstructured text. We use public API~(\url{clinicaltables.nlm.nih.gov}) to map them into ICD-10 codes. 
\item  \textbf{DrugBank}~\cite{wishart2018drugbank} (\url{www.drugbank.com}) is the database from which we obtain drug molecule structures in SMILES format - a concise ASCII string representation of chemical structures. 
\item \textbf{ZINC Molecule Database}~(\url{zinc.docking.org}) has $\sim$250K druglike molecules~\cite{sterling2015zinc,huang2021therapeutics} and is used to pretrain/train VAE model. 
\item \textbf{ICD-10 Coding System} (\url{www.icd10data.com}) is a medical classification list by the World Health Organization (WHO) \cite{anker2016welcome}. It catalogs codes for diseases, symptoms, and signs. In ICD-10, codes are structured hierarchically, with higher-level categories preceding more specific codes. 
\end{itemize}

\noindent\textbf{Data Split.} Then, we discuss the data split strategy in our paper, which is quite different from traditional machine learning methods that uses random split. 
AI models need to be evaluated on (future) unseen data. To simulate that setting, data split strategies are employed to partition the dataset into training, validation, and testing sets for unbiased evaluation of the machine learning models. In this paper, we leverage \textit{temporal split}, which refers to splitting the data samples based on their time stamps. The earlier data samples are used for training and validation, while the later data are used for testing. The reason is that the design of later clinical trials relies on earlier clinical trials. The training/test split ratio is 8:2. 

We use the clinical trials from 2018-2020 and 2021-2023 to create 2 test sets. We use all the trial samples before the time stamp of the test set to learn the model. If the drugs have been discovered and tested before, the data sample is categorized into the repurposing test set; otherwise, it is removed. Table~\ref{table:statistics} shows the data statistics.

\subsection{Baseline Methods} 
\label{sec:baseline} 

We consider the following baselines, \textbf{LR} (Logistic Regression), \textbf{RF} (Random Forest), \textbf{BPMF} (Bayesian Probabilistic Matrix Factorization) \cite{salakhutdinov2008bayesian,xiao2019bayesian,lu2019integrated}, \textbf{DeepDDI} (Deep Drug-Disease Interaction) \cite{ozturk2018deepdta,nguyen2019graphdta,chenlearning}.


\subsection{Evaluation Strategy and Metrics}.
\label{sec:metric}

When we make inferences at time $t$, we attempt to repurpose all the drugs in the existing drug database at the previous time stamp $t-1$. 
Two test sets are 2018-2020 and 2021-2023. 
When predicting at current time window, all the data before the time window are used for training. Each trial has one ground-truth drug. 
During inference, we infer a matching score for each drug in the drug database, given the disease code from each trial. 
The drug database contains 2,727 drugs by 2018, and 3,083 drugs by 2021. 
Then, we return the ranked drug list based on the matching scores. 
We consider \textbf{Hit@k\% accuracy} as the evaluation metric, which refers to whether the ground-truth drug appears in the first $k$\% positions of the ranked drug list. 


\subsection{Quantitative Results on Drug Repurposing}
\label{sec:result}
We demonstrate the experimental results in Table~\ref{table:repurposing}, we find that \mname consistently outperforms all baselines in all three metrics across two test sets. 
Compared with the best baseline BPMF, \mname obtains 16.5\% and 13.0\% relative improvement in the top@10\%, 11.7\% and 7.4\% absolute improvement on hit@30\% on 2018-2020 and 2021-2023 test sets, respectively. 
Also, we report the p-values to measure the statistical difference between \mname and the best baseline GraphDDI, the p-values are 0.018 and 0.013, less than the 0.05 threshold, which proves the statistical significance of our method over the best baseline. 
Both methods use the same neural architecture; the difference is \mname uses a contrastive learning framework to learn drug-disease relations.

\mname also significantly reduce the search space. 
The above metrics are challenging to achieve, because we have many repurposing candidates (a couple of thousands) and the ground-truth drug for a repurposing trial is just one single drug from this set. Our method achieves more than 90\% ``hit@30\%'' score in both test sets, which means for more than 90\% test data, our method can reduce 70\% search space.


\begin{table*}[h!]
\centering
\caption{Drug Repurposing results measured by hit@$k$ accuracy (higher the better) and ranking (lower the better). On 2018-2020 test set, the p-values between \mname and BPMF (the best baseline) are 0.00 and 0.00 on the two test sets, confirming its statistical significance. }
\label{table:repurposing}
\resizebox{1.41\columnwidth}{!}{
\begin{tabular}{lcccccc}
\toprule[0.6pt]
& \multicolumn{2}{c}{\bf 2018-2020 (2,727 drugs)} & \multicolumn{2}{c}{\bf 2021-2023 (3,083 drugs)} \\ 
Method & hit@10\% ($\uparrow$) & hit@30\% ($\uparrow$)  & hit@10\% ($\uparrow$) & hit@30\% ($\uparrow$) \\ \hline
LR & 42.4\% & 61.7\% & 38.5\% & 59.2\% \\ 
RF & 46.4\% & 70.8\% & 49.1\% & 76.3\% \\
BPMF & 49.6\% & 80.5\% & 57.7\% & 84.2\% \\
DeepDDI & 47.2\% & 76.6\% & 56.7\% & 81.3\% \\ 
\mname & \bf 57.8\% & \bf 92.2\% & \bf 65.2\% & \bf 91.6\% \\ 
\bottomrule[0.6pt]
\end{tabular}}
\end{table*}



\section{Conclusion}
\label{sec:conclusion}
In this paper, we have proposed DrugCLIP, a cutting-edge contrastive learning method for drug repurposing. 
Specifically, we formulate drug repurposing as a drug-disease interaction (a binary classification task), design representation learning to encode drug molecule and disease code, and use CLIP to align their representations without negative samples. 
Additionally, we have curated a drug repurposing dataset based on real-world public clinical trial records. We have conducted thorough experiments to confirm the superiority of the proposed DrugCLIP method.

\bibliographystyle{ACM-Reference-Format}
\balance
\bibliography{sample-base}

\end{document}